\documentclass[letterpaper, 10 pt, journal, twoside]{IEEEtran}
\usepackage[utf8]{inputenc}
\usepackage{amsmath}
\usepackage{amssymb}
\usepackage{graphicx}
\usepackage[style=ieee]{biblatex}
\usepackage{caption}
\usepackage{subcaption}
\usepackage{longtable}
\usepackage{supertabular}
\usepackage{upgreek}
\usepackage{gensymb}
\usepackage{textcomp}

\title{Improving Visual Feature Extraction in Glacial Environments}
\date{June 2019}
\bibliography{bib}
\AtBeginBibliography{\small}
\author{Steven D. Morad$^{1,2}$, Jeremy Nash$^{2}$, Shoya Higa$^{2}$, Russell Smith$^{2}$, Aaron Parness$^{2}$, and Kobus Barnard$^{3}$%
\thanks{Manuscript received: August 21, 2019; Revised October 27, 2019; Accepted November 19, 2019.}%
\thanks{This paper was recommended for publication by Editor Eric Marchand upon evaluation of the Associate Editor and Reviewers' comments.}
\thanks{A portion of this research was carried out at the Jet Propulsion Laboratory, California Institute of Technology, under a contract with the National Aeronautics and Space Administration.}
\thanks{$^{1}$Steven Morad is a graduate student in aerospace engineering at the University of Arizona, 1130 N Mountain Ave, Tucson AZ 85721, USA 
{\tt\small steven.morad@gmail.com}}
\thanks{$^{2}$Steven Morad, Jeremy Nash, Shoya Higa, Russell Smith, and Aaron Parness are employees at the Jet Propulsion Laboratory, California Institute of Technology, 4800 Oak Grove Dr, Pasadena, CA 91109, USA
{\tt\small \{jeremy.nash, shoya.higa, russel.g.smith, aaron.parness\}@jpl.nasa.gov} }
\thanks{$^{3}$Kobus Barnard is a professor of computer science at the University of Arizona, 1040 E 4th St, Tucson, AZ 85719, USA 
{\tt\small kobus@cs.arizona.edu}}
\thanks{Digital Object Identifier (DOI): see top of this page.}}%
\begin{document}
\maketitle
\markboth{IEEE Robotics and Automation Letters. Preprint Version. Accepted November, 2019}
{Morad \MakeLowercase{\textit{et al.}}: Improving Visual Feature Extraction in Glacial Environments}
\begin{abstract}
Glacial science could benefit tremendously from autonomous robots, but previous glacial robots have had perception issues in these colorless and featureless environments, specifically with visual feature extraction. This translates to failures in visual odometry and visual navigation. Glaciologists use near-infrared imagery to reveal the underlying heterogeneous spatial structure of snow and ice, and we theorize that this hidden near-infrared structure could produce more and higher quality features than available in visible light. We took a custom camera rig to Igloo Cave at Mt.~St.~Helens to test our theory. The camera rig contains two identical machine vision cameras, one which was outfitted with multiple filters to see only near-infrared light. We extracted features from short video clips taken inside Igloo Cave at Mt.~St.~Helens, using three popular feature extractors (FAST, SIFT, and SURF). We quantified the number of features and their quality for visual navigation by comparing the resulting orientation estimates to ground truth. Our main contribution is the use of NIR longpass filters to improve the quantity and quality of visual features in icy terrain, irrespective of the feature extractor used.
\end{abstract}


\section{Introduction}
\IEEEPARstart{S}{cientific} endeavors to many glaciers, such as Antarctica, are difficult and time-consuming. Extreme cold and lack of infrastructure restrict experiments. Some glaciers are littered with deadly crevasses, hidden under a deceiving layer of snow. Others break off or ``calve'' into the ocean, causing seismic events that register on the Richter scale. Glaciers are an environment ripe for automation.

Perception is a critical part of automation. Many machine vision algorithms rely on image features to extract meaning from an image. For navigation applications, these features are usually based on corners, regions in an image with large image gradients in two directions. Modern feature detectors find features that are invariant to camera translations and in-plane rotations. The motion of these features can inform a robot on where it is going or how the environment around it is changing -- an integral part of robotics.

\begin{figure}[t]
    \centering
    \begin{subfigure}{\linewidth}
        \includegraphics[width=\linewidth]{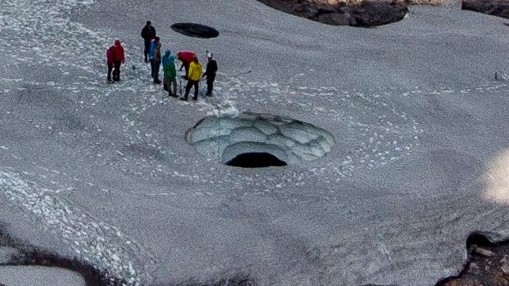}
        \subcaption{The ceiling entrance to Igloo Cave}
        \label{fig:cave_context}
    \end{subfigure}

    \begin{subfigure}{0.48\linewidth}
            \includegraphics[width=\linewidth]{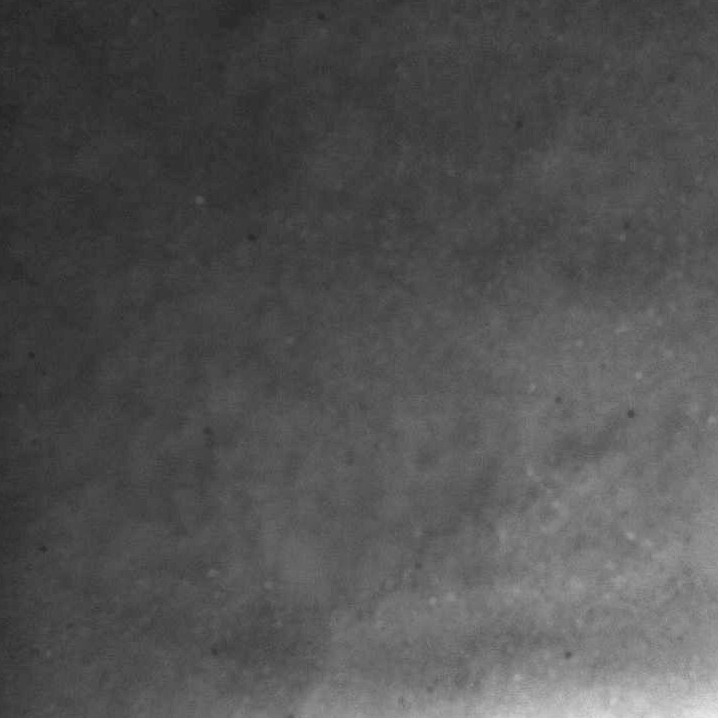}
            \subcaption{Left stereo picture without a filter}
            \label{fig:snow_vis_clahe}
    \end{subfigure}
    \begin{subfigure}{0.48\linewidth}
        \includegraphics[width=\linewidth]{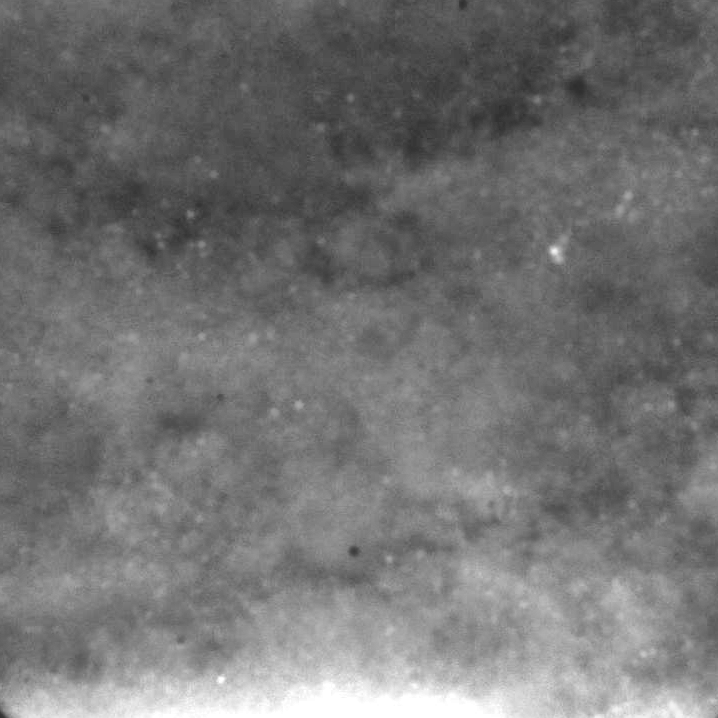}
        \subcaption{Right stereo picture with an 850nm filter (NIR-only)}
        \label{fig:snow_nir_clahe}
    \end{subfigure}
    \caption{Pictures from Igloo Cave at Mt.~St.~Helens. Both stereo pictures have contrast limited adaptive histogram equalization applied. The NIR-only image produces more and higher quality features.}
    \label{fig:fig_1}
    \vspace{-5mm}
\end{figure}

In our literature review, we found that a lack of visible features hamstrings robots in glacial environments. In many cases, successful glacial robots need to rely on other types of sensors. Featureless layers of snow and ice do not provide enough visual features for robotic decision making. However, glaciologists have tools to help them analyze snow and ice from afar. In particular, glaciologists make extensive use of near-infrared (NIR) light to differentiate between types of snow and ice. We leverage NIR light to improve the number and quality of visual features for machine vision applications. We investigate the optical properties of ice and snow to understand why glaciologists use this tool, and how we can adapt it for machine vision applications.

To test our hypothesis, we build a camera rig that detects both NIR and visible light, and use it to collect short video clips of Igloo Cave at Mt. St. Helens (Fig. \ref{fig:fig_1}). Igloo Cave is an ice cave that formed as the result of the volcanic activity of St. Helens. Our analysis of the video clips shows that filtered NIR vision generally outperforms unfiltered vision in feature extraction and camera orientation estimation in glacial environments such as Igloo Cave.

\section{Related Work}
\subsection{Glacial Robots and Vision}
The NASA funded Nomad robot was the first autonomous Antarctic robot. Its mission was to find meteorites in the Elephant Moraine. It was equipped with stereo cameras, but, as reported by Moorehead et al.~\cite{moorehead1999autonomous}: ``In all conditions, stereo [vision] was not able to produce sufficiently dense disparity maps to be useful for navigation'' . 

More recently, Paton et al.~\cite{paton2016dead} mounted stereo cameras on the MATS rover to explore the use of visual odometry in polar environments. They found that feature-based visual odometry performed poorly in icy environments: ``From harsh lighting conditions to deep snow, we show through a series of field trials that there remain serious issues with navigation in these environments, which must be addressed in order for long-term, vision-based navigation to succeed ... Snow is an especially difficult environment for vision-based systems as it is practically contrast free, causing a lack of visual features''.

Similar to Paton et al., Williams and Howard~\cite{williams2010developing} developed and tested a 3D orientation estimation algorithm on the Juneau Ice Field in Alaska. They wrote ``When dealing with arctic images, feature extraction is possibly the biggest challenge''. They used contrast limited adaptive histogram equalization (CLAHE) post-processing to enhance contrast and make features stand out better. Their algorithm can extract many more features than previously possible, but they still experience significant pose drift.

To summarize, previous attempts at glacial robots have had less-than-successful performance with vision in icy environments. By and large, this is mostly due to lack of visual features in vast sheets of ice and snow.
\subsection{Near-Infrared Filtering and Glaciology}
Near-infrared (750-2500nm) imaging is a known tool in glaciology. Champollion used NIR imaging to get better images of hoarfrost in Antarctica \cite{champollion2013hoar}. NIR imagery from the MODIS satellite has been used to calculate continent-wide surface morphology and ice grain size measurements in Antarctica \cite{scambos2007modis}. Matzl and Schneebeli took NIR photographs of roughly one square meter of ice and snow, generating a 1D spatial map of grain structure within the snowpack~\cite{matzl2006measuring}. They found found that at meter-scales, differences in the snowpack are visible in NIR .
\subsection{Near-Infrared Feature Extraction}
Relatively little work on feature extraction has been done in the near-infrared. Kachurka et al.~ \cite{kachurka2019swir} evaluated standard ORB SLAM in the short-wave IR (SWIR), with the addition of a small keyframe modification to reduce the occurrence of reinitialization. Johannsen et al.\cite{johansson2016evaluation} suggest that the ORB feature extractor performed best in their thermal IR feature extractor benchmark . Neither of these evaluate performance in the NIR waveband. Additionally, glacial environments appear drastically different than their urban test environments. Sima and Buckley~\cite{sima2013optimizing} and Ricaurte at al.~\cite{ricaurte2014feature} discuss optimizing feature extractors in SWIR and thermal IR to enable matching to features captured in visible light, but again, not for icy evironments. 
\section{Method}
\subsection{Scattering Models}
\begin{figure}
    \centering
    \includegraphics[width=\linewidth]{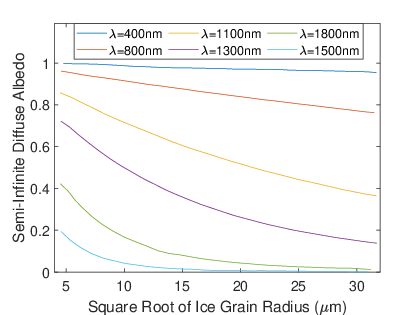}
    \caption{NIR albedo depends much more on ice grain size than visible light. For reference, the human eye is most receptive at 0.56$\mu$m \cite{bradley1991achromatizing}. Adapted from \cite{wiscombe1980model}.}
    \label{fig:albedo}
    \vspace{-5mm}
\end{figure}

Wiscombe's seminal work on the optics of snow and ice utilizes Mie theory to describe scattering. Their model describes the optics of ice and snow from 300nm to 5000nm. They find that the reflectance of ice grains between 750 and 1400nm is mostly dependent on the size of the grains \cite{wiscombe1980model} (Fig. \ref{fig:albedo}), thereby exposing structure invisible outside those wavelengths. For reference, visible light ends at 740nm. Since their work was published, several other papers have confirmed that snow albedo (brightness) is sensitive to ice grain size in NIR wavelengths \cite{kokhanovsky2004scattering} \cite{xie2006effect}.
\subsection{Specific Surface Area and Grain Size}
Ice and snow are made up of small ice crystals called ice grains that measure from tens to thousands of microns across \cite{scambos2007modis}. The term ``grain size'' refers to the diameter of these grains, but is sometimes misleading. In optics, the grain size of ice has two meanings: the true size of the grain or the optical size of the grain. Thus far, we have referred to the optical grain size. The optical size is used in idealized lighting models to reconcile the error between modeled and observed values for a specific true grain size.

The specific surface area (SSA) of snow and ice is defined as the ratio between the surface area and volume of the ice. SSA is strongly coupled with optical grain size \cite{mitchell2002effective}, but can also effectively represent differences in grain shape. SSA has been shown to better represent the optical bulk-properties of real-world snow and ice \cite{grenfell1999representation}. The SSA can also represent spatially varying properties of snow and ice, such as air content or ice age \cite{legagneux2002measurement}. While individual ice grains are usually too small to resolve by camera, regions of snow and ice with differing SSA are not. Varying SSA regions appear differently when viewed in NIR light. These differences in NIR light produce more numerous and distinct visual features than if viewed in visible light.
\section{Experiment}
We set out to compare the number and quality of features extracted from NIR and visible light imagery. First, we define the scenes where video is taken. Then, we discuss the camera rig design and camera parameters. We go over the video capture procedure and the metrics we use to evaluate each scene.
\subsection{Cave Scenes}
We analyze video from four different scenes inside Igloo Cave at Mt.~St.~Helens. The first scene is a featureless firn wall, the second scene is a striated firn wall, and the third scene is planar snow. The fourth scene is a walking tour around one portion of the cave. Indirect sunlight illuminates all but the planar snow scene, which is illuminated by the lamp on the camera rig. 
\subsection{Camera Rig Design}
\begin{figure}
    \centering
    \includegraphics[width=\linewidth]{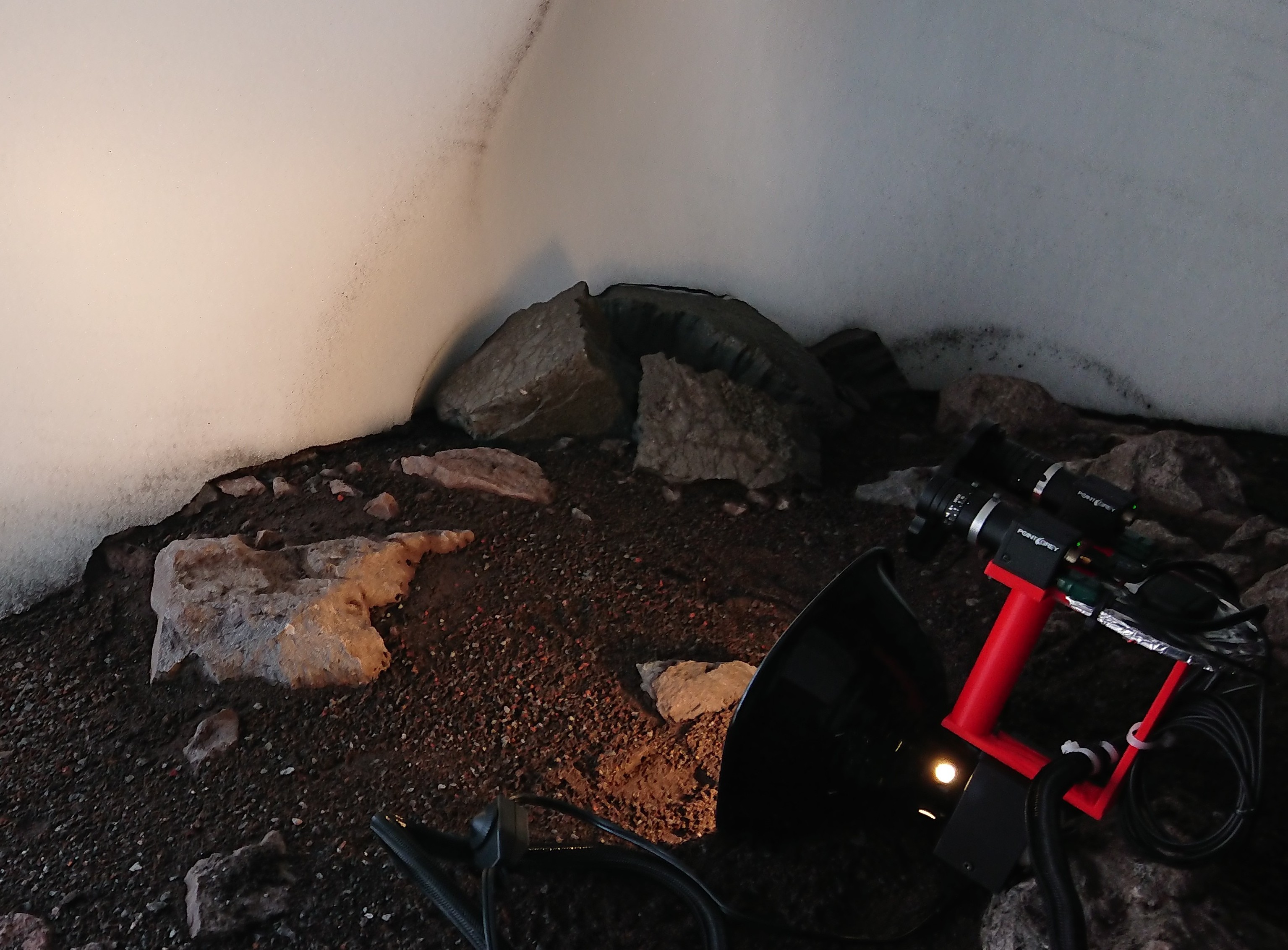}
    \caption{The camera rig and a firn wall at St. Helens}
    \label{fig:example}
    \vspace{-5mm}
\end{figure}

A hand-held camera rig was built to collect NIR data and compare it to visible light. We mount two identical PointGrey FLEA-3 monochrome cameras to a 3D printed structure in a stereo configuration with a 10cm baseline (Fig. \ref{fig:example}). The right camera has a filter wheel flush with the lens assembly. The filter wheel contains five NIR longpass filters with cut-on wavelengths of 800nm, 850nm, 900nm, 950nm, and 1000nm. These filters block light below their cut-on wavelength. We also attach a terrarium lamp on the underside of the rig, centered between the two cameras. The terrarium lamp has a ceramic reflector that reflects light in both visible and IR spectrums. A 75W halogen-tungsten incandescent bulb sits in the terrarium lamp to provide smooth, continuous illumination over both the visible and infrared spectrums. Mounted between the cameras is a VectorNav VN-200 inertial measurement unit (IMU) that provides ground truth orientation data. The VN-200 provides yaw to within 0.3\textdegree{} and pitch/roll to within 0.1\textdegree{} RMS, and runs at 800Hz.
\subsection{Camera Parameters}
Varying lighting conditions and the differing transmissivity of each filter made hand-setting camera parameters for each scene very difficult. Due to the significant difference in light received by the sensors, one set of parameters would not work for both cameras. By setting camera parameters differently for each camera, we could bias the results. For these reasons, we set the cameras to auto mode. Auto mode automatically sets the analog gain, shutter speed, and sharpness of each camera. Because the NIR camera receives less light, it has a higher gain and prolonged exposure, which results in noisier and blurrier video. This provides some advantage to the visible light camera, but we did not attempt to quantify the extent of the advantage. We set the camera to capture 20 frames per second, but due to in-situ video compression the framerate would sometimes drop as low as 15 frames per second.
\subsection{Procedure}
We hold the camera rig by hand and take short videos while trying to keep the rig from moving too much. In all scenes, the rig is between one and six feet from the region of interest. If the scene is too dark for the unfiltered camera, the illuminator is turned on. For each scene, we run the experiment five times, each time cycling to the next NIR longpass filter on the right camera. For the cave tour, the camera rig is held a few feet from the cave wall as the operator walks about the cave. The path is identical for all filters. In our videos, we observe only snow and ice. Special care is taken to ensure that no rocks or foliage appear in any of the videos. Videos that contained enough volcanic ash to affect the results were discarded, except for the cave tour. 

\subsection{Preprocessing}
Each image frame goes through a preprocessing pipeline before analysis. Lens distortion causes straight lines to appear slightly curved in the image; images are rectified to remove this effect.
Next, we remove vignetting created by the filter wheel. Hough circles are used to detect the vignette perimeter. Once the perimeter is determined, we inscribe a bounding square in the hough circle. On both cameras, we only use data within the bounding square. 
Finally, the resulting image goes through CLAHE to improve contrast, as Williams and Howard suggest for icy environments \cite{pizer1987adaptive} \cite{williams2010developing}.

\subsection{Metrics}
We evaluate multiple feature detectors: SIFT \cite{lowe1999object}, SURF \cite{bay2006surf}, and the slightly modified scale-space version of FAST used in the ORB paper \cite{mur2015orb}. All feature detectors we use are scale-invariant by way of a scale-space pyramid. Each feature detector, except for SURF, uses default OpenCV parameters to reduce the chance of biasing parameters to improve NIR imagery at the expense of visible light imagery. The minimum Hessian threshold for SURF is raised to 500 to produce features similar in quantity and quality to SIFT and FAST.

\subsubsection{Feature Count}
The most straightforward metric is counting the number of features in each picture. Five features is the practical lower bound for visual pose estimation \cite{nister2004efficient}. With RANSAC, more features result in more samples for pose estimation at the expense of some computational overhead \cite{nister2005preemptive}. We take the median number of features per frame over the entire video. Then, we take the mean over all feature extractors.

\subsubsection{Valid Orientation Percentage (VOP)}
Just counting the raw number of extracted features can be misleading because ``false features'' are counted. False features are features created from camera noise or other sources that do not persist between frames and are not useful for vision. The ultimate test for feature extraction is whether the features are good enough to provide valid visual odometry estimates.

We estimate the essential matrix $\mathbf E$ using our lab-estimated intrinsics $\mathbf K$. We feed the keypoints from a specific extractor to OpenCV's \texttt{findEssentialMat} function which uses Nister's five-point algorithm to determine $\mathbf E$ \cite{nister2004efficient} \cite{opencv_library}. Once we have $\mathbf E$, we decompose it into a rotation matrix and translation vector. Due to the inherent noisiness caused by double integration of accelerometer data, we cannot analyze translation. We compare the relative difference in orientation between two frames to the ground truth value recorded by the IMU. If the relative 3D rotation is within five degrees of the ground truth, we consider the estimate valid and invalid otherwise. If for any reason we are unable to construct or decompose $\mathbf E$, we consider that estimate invalid. We provide the percentage of frame pairs with a valid estimate out of the total number of frame pairs. In other words, this metric describes how often we are are able to accurately estimate relative camera orientation from the extracted features. We call this metric the Valid Orientation Percentage (VOP).

\section{Results}
Although we used filters up to 1000nm, indirect lighting conditions combined with reduced camera sensitivity results in pitch black videos for longer wavelength filters. Even with the lamp, some filter and scene combinations were too dark for analysis. For this reason, we exclude the 950nm and 1000nm filter results.



\begin{figure}[t]
    \centering
    \includegraphics[width=\linewidth]{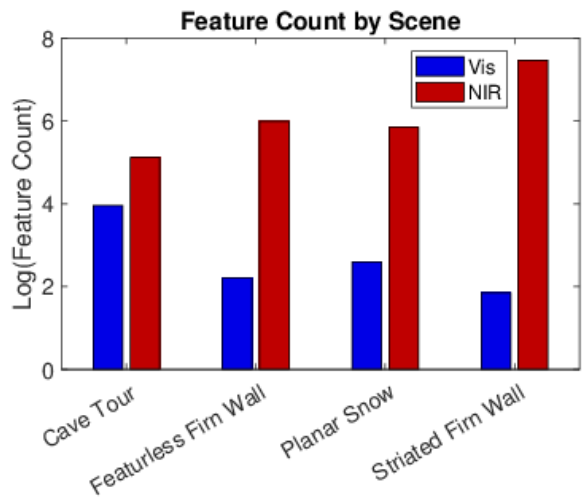}
    \caption{Natural log of the feature count, broken down by scene. NIR imagery produced orders of magnitude more features.}
    \label{fig:feat}
    \vspace{-5mm}
\end{figure}
\begin{figure}[th!]
    \centering
    \begin{subfigure}{\linewidth}
        \includegraphics[width=\linewidth]{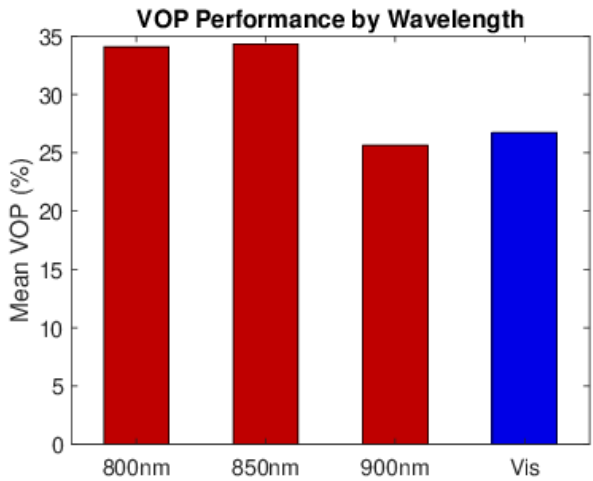}
        \subcaption{Results breakdown by filter wavelength}
        \label{fig:wavelength_perf}
    \end{subfigure}
    \begin{subfigure}{\linewidth}
            \includegraphics[width=\linewidth]{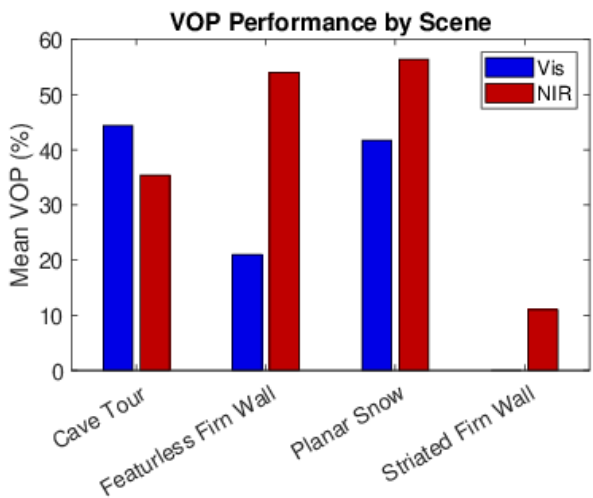}
            \subcaption{Results breakdown by scene}
            \label{fig:plot}
    \end{subfigure}
    \begin{subfigure}{\linewidth}
        \includegraphics[width=\linewidth]{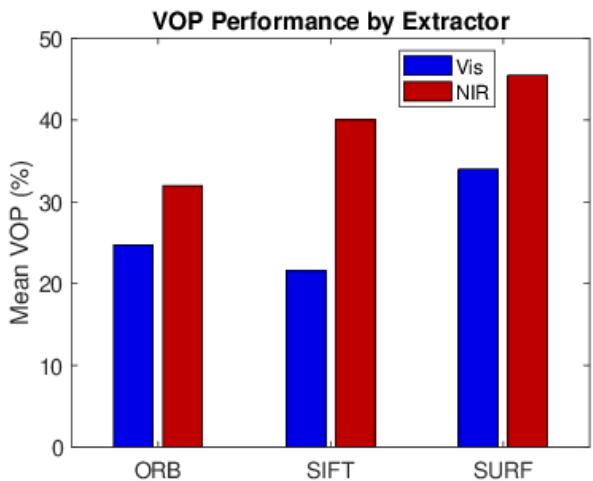}
        \subcaption{Results breakdown by feature extractor}
        \label{fig:extractor}
    \end{subfigure}
    \caption{Comparisons of orientation validity broken down into various buckets. For Fig. \ref{fig:plot}, \ref{fig:extractor}, NIR corresponds to the best performing filter for each scene.}
    \label{fig:results}
\end{figure}

We provide our results in Fig. \ref{fig:feat}, \ref{fig:results}. CLAHE modified imagery always outperformed non-CLAHE imagery, so we omit the non-CLAHE results. The overall best performing filter is 850nm, beating visible light (no filter). The 800nm filter performed almost as well as the 850nm filter, and still beat unfiltered light. Filtered light outperformed unfiltered light except in the cave tour scene, due to volcanic ash that provided additional features in the visible spectrum. Looking at performance arranged by feature extractor, filtered light outperformed unfiltered light for all tested feature extractors (Fig. \ref{fig:extractor}). This shows that the NIR performance gains are extractor agnostic -- NIR provides more visual information for the feature extraction and visual odometry problems, irrespective of the extractor used.

\begin{figure}
    \centering
    \includegraphics[width=\linewidth]{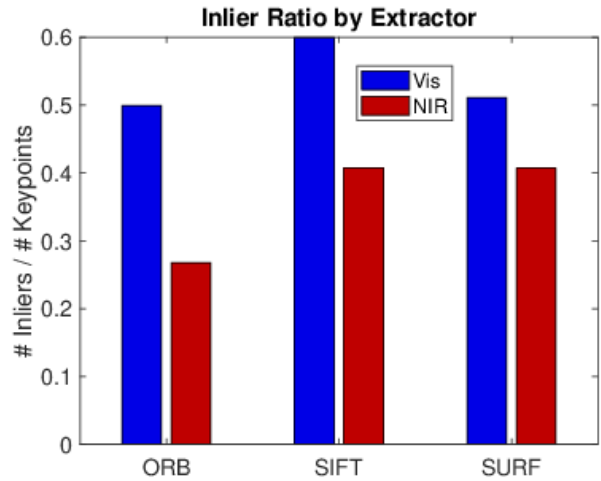}
    \caption{Ratio of inliers to total keypoints (inliers plus outliers), broken down by extractor. NIR corresponds to the best performing (VOP) filter for each scene. The ratio only considers frames where the essential matrix is found.}
    \label{fig:ratio}
\end{figure}



\section{Discussion}

\begin{figure}[]
    \centering
    
    \begin{subfigure}{0.48\linewidth}
            \includegraphics[width=\linewidth]{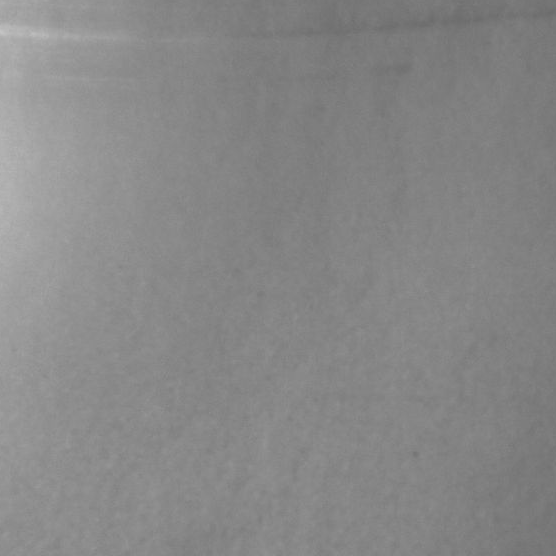}
            \subcaption{Left stereo picture without a filter}
            \label{fig:striated_vis}
    \end{subfigure}
    \begin{subfigure}{0.48\linewidth}
        \includegraphics[width=\linewidth]{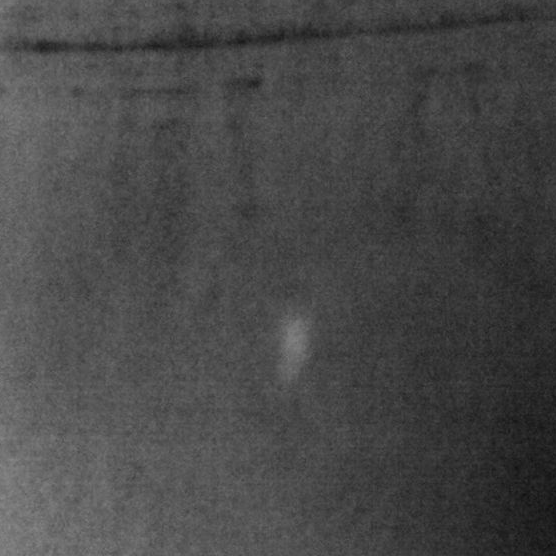}
        \subcaption{Right stereo picture with an 850nm filter}
        \label{fig:striated_nir}
    \end{subfigure}
    \caption{Stereo imagery of a striated firn wall. The melt-freeze crust near the top of the stereo images provides features in NIR.}
    \label{fig:striated}
    \vspace{-5mm}
\end{figure}


\subsection{Concrete Examples}
We attempt to connect the results back to our SSA hypothesis through qualitative means. The planar snow scene is the best example of the spatially-varying SSA. When comparing the visible light image (Fig. \ref{fig:snow_vis_clahe}) to the NIR image (Fig. \ref{fig:snow_nir_clahe}), there is a stark difference. The NIR image almost looks like a cloudy sky or a nebula. The darker regions are those with smaller SSA. These are likely regions of older snow, where dendritic grains transition to round grains \cite{smith2009tracking}. The brighter areas could be regions of new snow with higher SSA. 

Also visibly interesting is the striated firn wall scene (Fig. \ref{fig:striated}). The striation in this scene is known as melt-freeze crust, where melting snow or rain creates a layer of water, then refreezes producing large ice grains \cite{smith2008observation}. These large ice grains result in a small SSA and a dark streak in the NIR image (Fig. \ref{fig:striated_nir}). Note that in the unfiltered image, the SSA has little effect and the streak is barely visible (Fig. \ref{fig:striated_vis}).

\subsection{Practical Considerations}
While other light spectrums have interesting interactions with ice crystals, NIR light is the most practical. Most silicon CMOS and CCD camera sensors are sensitive to NIR light. Many machine vision cameras come without a NIR-blocking filter, allowing them to view NIR light out of the box. Consumer cameras tend to have NIR blocking filters to restrict the sensor to the human vision range. These filters can easily be replaced with NIR longpass filters, allowing almost any commercial camera to see in only NIR wavelengths.

While cameras sensitivities vary, the spectral sensitivity of the Flea3 cameras is representative of other commercial cameras. For most cameras, we expect that 800nm and 850nm pass filters with CLAHE post-processing will produce the best visual features. The noisy low-light photography produced by the 900nm and higher filters combined with noise-sensitive CLAHE results in many features created from noise. A sensor that is more sensitive to NIR light would perform better in longer wavelengths with CLAHE. Most of the testing occurred inside a darkened cave, the darker filters will likely perform better outside in direct sunlight.

\subsection{Future Work}
The cameras we used only touch the very beginning of the NIR spectrum. With specialized NIR sensors, it may be possible to extract even more features. Indium-Gallium-Arsenide sensors are commercially available and span the full NIR spectrum. Furthermore, other types of sensors such a polarization sensors may provide additional benefits.

We evaluated the feature extractors without changing the extractor default parameters to isolate light wavelength as the independent variable. We have shown that NIR generally outperforms visible light in this task, the next step is to find the optimal feature extractor and associated parameters.

All analyzed scenes are from inside Igloo Cave at St. Helens, which means that all imagery is ``indoors''. While this is important for NASA's future goals, future research should strive to obtain test data from outdoor environments to test far-field visual navigation as well.

Although we quantified orientation error between frames, we did not explore how NIR imagery would perform long-term in a SLAM scenario. We would have liked to test this, but we had no way to correct IMU drift while collecting ground truth data. Future experiments should focus on improving the quality of ground truth data.

\section{Conclusion}
Our experimental results from Igloo Cave suggest that NIR light is an attractive alternative to visible light for feature extraction and visual navigation in glacial environments. In most of our cases, the NIR imagery outperformed visible light imagery. We were able to accurately estimate camera orientation much more often in NIR imagery than in visible light imagery. The biggest disadvantage of using NIR pass filters on a regular camera indoors is the reduced amount of light that hits the sensor. Longer exposures and higher gain can mitigate this this to an extent, but indoors, ensuring adequate lighting is very important. Above 850nm, the light reduction started to severely impact the image quality in the form of blur or noise. With larger illuminators or more sensitive cameras, it is likely that the optimal wavelength will be higher and perhaps the performance even better.

\printbibliography
\end{document}